%% file: main.tex
\documentclass[11pt]{article}

\usepackage[preprint]{acl}
\usepackage{xurl}
\usepackage{float}
\usepackage{tikz}
\usepackage{caption}
\usetikzlibrary{arrows.meta, positioning}

\usepackage{common}
\usepackage{siunitx}
\usepackage[T1]{fontenc}
\usepackage[utf8]{inputenc}
\usepackage{times}
\usepackage{latexsym}
\usepackage{microtype}

\newcommand{\skillmd}{\texttt{SKILL.md}\xspace}
\newcommand{\loongdoc}{\textsc{LoongDoc}\xspace}
\newcommand{\infbench}{\ensuremath{\infty}Bench\xspace}

\newcommand{\rawdoc}{\texttt{raw}\xspace}
\newcommand{\flatdisc}{\texttt{flat}\xspace}
\newcommand{\hierdisc}{\texttt{hierarchical}\xspace}
\newcommand{\hybridrag}{\texttt{hybrid-rag}\xspace}
\newcommand{\codex}{Codex\xspace}
\newcommand{\piagent}{Pi\xspace}
\newcommand{\claudecode}{Claude-Code\xspace}
\newcommand{\gptmini}{\texttt{gpt-5.4-mini}\xspace}
\newcommand{\qwenmodel}{\texttt{qwen3.6-27b}\xspace}
\newcommand{\haikumodel}{\texttt{claude-haiku-4.5}\xspace}
\newcommand{\enmc}{\texttt{En.MC}\xspace}
\newcommand{\enqa}{\texttt{En.QA}\xspace}
\newcommand{\zhqa}{\texttt{Zh.QA}\xspace}
\newcommand{\concurrenttask}{\texttt{concurrent\_task=1}\xspace}

\definecolor{rowraw}{HTML}{F4F4F4}
\definecolor{rowflat}{HTML}{DCE9F7}
\definecolor{rowhier}{HTML}{FCE8D5}
\definecolor{rowrag}{HTML}{E8DFF1}

\title{Is Progressive Disclosure All You Need for Long-Context Agents?}

\author{
    Yifeng He\textsuperscript{\rm 1},\,
    Yinzhe Zhao\textsuperscript{\rm 2}\thanks{Work done while interning at UC Davis.},\,
    Jicheng Wang\textsuperscript{\rm 1},\,
    Hao Chen\textsuperscript{\rm 3} \\
    \textsuperscript{\rm 1}University of California, Davis\\
    \textsuperscript{\rm 2}Zhejiang University,\,
    \textsuperscript{\rm 3}The University of Hong Kong\\
    \texttt{yfhe.cs@gmail.com,\, chenho@hku.hk}
}

\begin{document}

\maketitle

\input{src/abs.tex}
\input{src/intro.tex}
\input{src/related.tex}
\input{src/method.tex}
\input{src/environment.tex}
\input{src/results.tex}
\input{src/conclusion.tex}

\input{src/limitations.tex}
\input{src/ethics.tex}

\section*{Acknowledgments}
This material is based upon work supported by UC Noyce Initiative.

\bibliography{main}

\input{src/appendix.tex}

\end{document}

%% file: src/abs.tex
\begin{abstract}
Long-document question answering usually forces a choice between loading the whole document into the context window and bolting on a separate retriever.
Agentic AI suggests a broader option, giving the agent the document path and letting it decide how and what to read.
Agent Skills, a standard for packaging expertise into folders an agent loads on demand,
supply a ready mechanism: \emph{progressive disclosure}, which exposes only what a query needs, from a short description down to the specific passages.
Practitioners rapidly adopted this pattern for book-length understanding tasks,
but the evidence for it has been anecdotal.
We run the first controlled study of the pattern, comparing raw-document navigation and several designs of Agent Skills packs against a classical hybrid retriever across three agent harnesses and three model families on \infbench{}.
On a single book, the gain depends on the harness, running large when the agent navigates the raw document poorly but near zero when a strong agent harness already locates and reads the right passages on its own.
When scaling up to tasks that span many books,
raw-document navigation collapses while one-level progressive disclosure degrades more slowly and pulls ahead.
A second, deeper routing level never helps and sometimes breaks accuracy outright, so one level is enough.
Progressive disclosure buys context, not intelligence: 
it is redundant while a strong agent can locate the right passages itself, and decisive once the corpus grows too large to navigate by reading.
\end{abstract}

%% file: src/intro.tex
\section{Introduction}
\label{sec:intro}

Long-document question answering (QA) forces a choice between two well-studied options: load the entire document into the context window and lean on the model's long-context attention, or retrieve a small set of chunks with an external retriever.
Neither option wins outright~\citep{li2024lcorrag,yu2025lcvsrag}, and on long inputs the effective context size falls well short of the advertised window~\citep{liu2023lostinthemiddle,hsieh2024ruler}.
Agentic AI reframes the choice: give the agent the document and let it decide what to read and when.
Claude Code made exactly this move, dropping the retrieval index its early versions used in favor of letting the agent search the code base on demand~\citep{cherny2025claudecode}.
But its engineers report the switch from experience, not a controlled comparison.

Agent Skills turn that option into a mechanism.
The standard packages expertise as a directory of instructions and resources~\citep{anthropic2025skills,xu2026agentskillssurvey}: root the document at a \skillmd{} file, hand it to the agent, and let \emph{progressive disclosure} do the retriever's job, keeping a short description of each skill always in context so the agent reads a skill's body and its bundled files only when that description matches the task.
The pattern spread on engineering intuition long before anyone measured it~\citep{hn2025skillslaunch,hn2025willisonskills}.
For book-length material, it is urged as a way to package a whole book so that an agent reads only the chapter a question needs~\citep{virgiliojr2026booktoskill,praison2026booktoskill,pyshine2026booktoskill}.
However, whether it should displace retrieval is itself contested~\citep{ewerlof2026ragskill}, and no controlled study has yet tested it with an agent on a standardized long-document benchmark.

This practice raises two open empirical questions.
First, \emph{does progressive disclosure work for book-length understanding?}
Second, \emph{what is the best way to structure the skill pack?}
Both are sharper than they first appear, because the Agent Skills specification bundles two design choices the practitioner literature never separates: how deep the disclosure recurses, and where the per-chunk index physically lives.
We pull them apart by comparing the three approaches an agent can take to a book-length document.
Raw-document navigation (\rawdoc) gives the agent the document with no skill pack.
Flat disclosure (\flatdisc, \autoref{fig:flat}) exposes it as a single \skillmd{} whose always-loaded description routes the agent to the book and whose body indexes chunk files under \texttt{references/} that it reads on demand.
Hierarchical disclosure (\hierdisc, \autoref{fig:hier}) pushes the same idea one level further: each chunk becomes its own skill with an always-loaded description, and a meta-router routes among them.
Both are progressive disclosure in the sense of~\citet{anthropic2025skills}; they differ only in routing \emph{depth} and in whether per-chunk descriptions pay always-loaded (\hierdisc) or load-on-activation (\flatdisc) context.

We build both disclosure packs by implementing the widely adopted \emph{book-to-skill} recipe~\citep{virgiliojr2026booktoskill}: split a book along its own structure into chunks, then attach a short LLM-written description to each; both packs share one chunk set, so the only thing that varies is how the agent reaches a passage.
To run the comparison, we turn \infbench{}~\citep{zhang2024infinitebench} into an agent environment,
following the practice of established prior evaluation work~\citep{merrill2026terminalbench,li2026skillsbench},
and we test across three agent harnesses and three model families, with \hybridrag, a classical retrieve-and-rerank baseline, as an external reference.

Our findings are three-fold.
On a single book, flat disclosure helps only to the extent that the agent cannot already navigate the document: it matches or exceeds raw-document navigation under \piagent and \claudecode, but adds nothing under \codex, whose bare model already greps for the entities each question names.
The hierarchical pack never beats the flat one, and sometimes collapses accuracy outright, dropping \enmc from $0.9126$ to $0.6398$ on \piagent \gptmini.
At library scale the picture flips: bundling twenty books sinks even \codex under raw navigation, where English open QA falls to $0.26$, while flat disclosure holds it at $0.46$.
The same scaling is an efficiency loss, not only an accuracy loss: as the corpus grows, accuracy falls and the per-question cost of reaching any fixed accuracy rises, so a larger library is strictly less efficient to read.
Progressive disclosure thus buys context, not intelligence: its gain is negligible for a single book under a strong agent but decisive for a library, and one level is enough.
Our contributions are:
\begin{enumerate}
    \item \textbf{A controlled comparison of three agentic reading approaches (\autoref{sec:method}).} We cast raw-document navigation, flat disclosure, and hierarchical disclosure as three routes to the same book, building the two disclosure packs over one fixed chunk set so that only the wiring varies.
    \item \textbf{A long-context agent environment (\autoref{sec:env}).} We turn static book QA into an interactive environment that logs the agent's trajectory and emits a per-run reward, a reproducible testbed for long-context agents.
    \item \textbf{The first systematic study of progressive disclosure for long-document understanding (\autoref{sec:results}).} We measure a pattern practitioners argue at scale but have never validated, on \infbench{} across three harnesses and three model families.
\end{enumerate}

%% file: src/related.tex
\section{Related work}
\label{sec:related}

\subsection{Reading long documents}

\paragraph{Long-context modeling}
Bigger context windows have not bought uniformly better long-context use: \citet{liu2023lostinthemiddle} document U-shaped performance over context position, and \citet{hsieh2024ruler} show effective context size lagging far behind the advertised window.
We evaluate on \infbench{}~\citep{zhang2024infinitebench}, a book-length benchmark~\citep{bai2024longbench,bai2024longbenchv2,wang2024novelqa,karpinska2024nocha,kocisky2018narrativeqa} where frontier models degrade heavily, so an agentic alternative earns a controlled test.

\paragraph{Retrieval augmented generation}
Retrieval augmented generation (RAG) softens raw long-context degradation but stays unstable across tasks: long context wins when resources permit while retrieval wins on cost~\citep{li2024lcorrag}, and the recent literature reaches contradictory conclusions~\citep{yu2025lcvsrag}.
Hierarchical retrieval augments flat chunk-and-rank with tree summaries, cascading metadata, and entity graphs~\citep{sarthi2024raptor,chen2024hiqa,edge2024graphrag}, improving coverage on book-length and larger inputs but locking the structure inside the retriever.
We instead expose the same hierarchy as an artifact the agent reads and navigates directly.

\paragraph{Agentic reading}
A parallel line lets the model navigate the document directly: recursive summary trees~\citep{chen2023memwalker}, gist-based pagination with lookup on demand~\citep{lee2024readagent}, typed query-decomposition plans~\citep{sun2023pearl}, and multi-agent splits coordinated across workers~\citep{zhao2024longagent,zhang2024chainofagents}.
All of these build their navigation structure \emph{at run time} and treat it as transient, not a persistent artifact, and most evaluate on long-document reading-comprehension benchmarks such as QuALITY and NarrativeQA.
The closest neighbor that builds its structure offline is Corpus2Skill~\citep{sun2026navigate}, which distills an enterprise corpus into a hierarchical skill directory the agent navigates top-down.
Its generalization study is telling: navigation beats flat retrieval on single-domain corpora with a recoverable taxonomy but not on open-domain factoid pools, exactly the regime distinction that favors a single structured long document.
We depart from this run-time line on every axis: the skill pack is built once from document structure and reused across queries, the routing index is persistent and human-inspectable, and the navigation interface is the file system, not a bespoke tree API.

\subsection{Agent skills}

The agent-engineering community has converged on a complementary pattern, surveyed and formalized around progressive disclosure~\citep{xu2026agentskillssurvey,zhou2026externalizationllmagentsunified}: package a domain or document as a directory rooted at a \texttt{SKILL.md} routing file whose when-to-use rules the agent consults before it reads any content~\citep{anthropic2025skills,anthropic2025context,virgiliojr2026booktoskill,karaaslan2026skillseekers}.
Such skills are hand-curated, induced from agent trajectories~\citep{wang2026skillinduction}, or, as here, derived from a document's own structure.
Adopted at scale on engineering intuition before anyone measured it, the pattern is promoted as a near-replacement for retrieval~\citep{willison2025skills,thoughtworks2026disclosure} and, for book-length material, as a way to package whole books and corpora as on-demand knowledge bases~\citep{virgiliojr2026booktoskill,praison2026booktoskill,pyshine2026booktoskill,bakal2026knowledgeactivationaiskills}, with token-savings claims reported under no accuracy control~\citep{pocock2026skills}.

Measurement has now begun, and two of its findings bear directly on how we build a skill pack.
First, curated skills help but model-authored ones do not: \textsc{SkillsBench}~\citep{li2026skillsbench} reports that curated packs lift performance while model-authored packs add nothing, and \citet{huang2026llmauthored} trace the null to every section of an \mbox{LLM}-written skill, which argues for fixing a skill's structure from the document rather than letting the model author it wholesale.
Second, the routing metadata carries the weight: auditing \num{138000} public \skillmd{} files, \citet{zhang2026skillmd} show that valid metadata is what makes a skill reliably retrievable.
Yet all of this measures skill uplift against a no-skill baseline on heterogeneous agentic tasks.
None tests the pattern against a raw-document baseline on a standardized long-document benchmark, or isolates which design axis (routing depth, chunk granularity, index location) carries the effect.
We close that gap.

%% file: src/method.tex
\section{Agentic long-context reasoning}
\label{sec:method}

Agentic long-context reasoning hands the whole document to an agent and asks it to answer questions about it.
A book does not fit in a context window, so the agent must first find the passage that holds the answer, and practitioners wire that search three ways without comparing them on equal footing.
The first hands the agent the raw document and lets it navigate on its own (\rawdoc).
The second and third package the document as an Agent Skills~\citep{anthropic2025skills} pack and let progressive disclosure route the agent to the passage, at one of two depths: a single flat index (\flatdisc) or a recursive hierarchy of per-chunk skills (\hierdisc).
We build the two disclosure packs by implementing \emph{book-to-skill}~\citep{virgiliojr2026booktoskill}, a widely adopted open recipe for turning a long document into a skill pack, and run all three approaches against the same book.
Both packs share a single chunk set (\autoref{sec:method:pipeline}), so the disclosure wiring is the only variable (\autoref{sec:method:isolation}).

\subsection{The book-to-skill pipeline}
\label{sec:method:pipeline}

We run the following steps once per book, so the chunk set and the descriptions are fixed inputs to both disclosure approaches, not variables.

\paragraph{Chunking}
We split the book along its own structure, cutting one chunk at each chapter heading and following that division whenever the book provides it.
When a book exposes no such headings, we fall back to fixed-size chunks of about \num{4000} words, split at paragraph boundaries.
Each chunk becomes a file under \texttt{references/}, addressable by name.

\paragraph{Description}
For each chunk we prompt an LLM for a short description of what the chunk covers, paired with the named entities it introduces, and we prompt it once more over those chunk descriptions for a book-level description (\appautoref{sec:appendix:prompts}).
These descriptions are the routing metadata the agent reads before it opens any chunk, the one place where \citet{zhang2026skillmd} found metadata to be load-bearing for retrievability.

\subsection{The three approaches}
\label{sec:method:approaches}

\begin{figure}[t]
    \centering
    \includegraphics[width=\columnwidth]{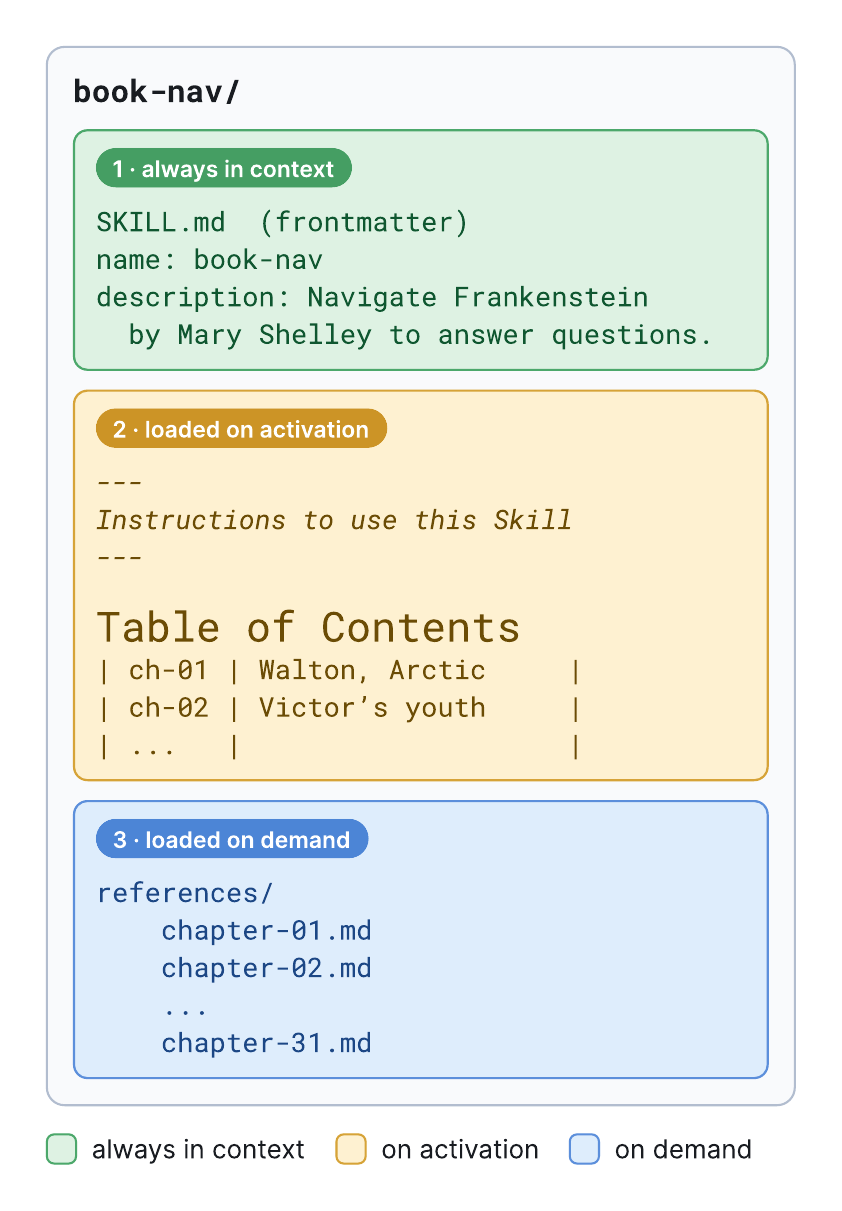}
    \caption{%
        The \flatdisc pack. %
        Example is \emph{Frankenstein}.
    }
    \label{fig:flat}
\end{figure}

\begin{figure}[t]
    \centering
    \includegraphics[width=\columnwidth]{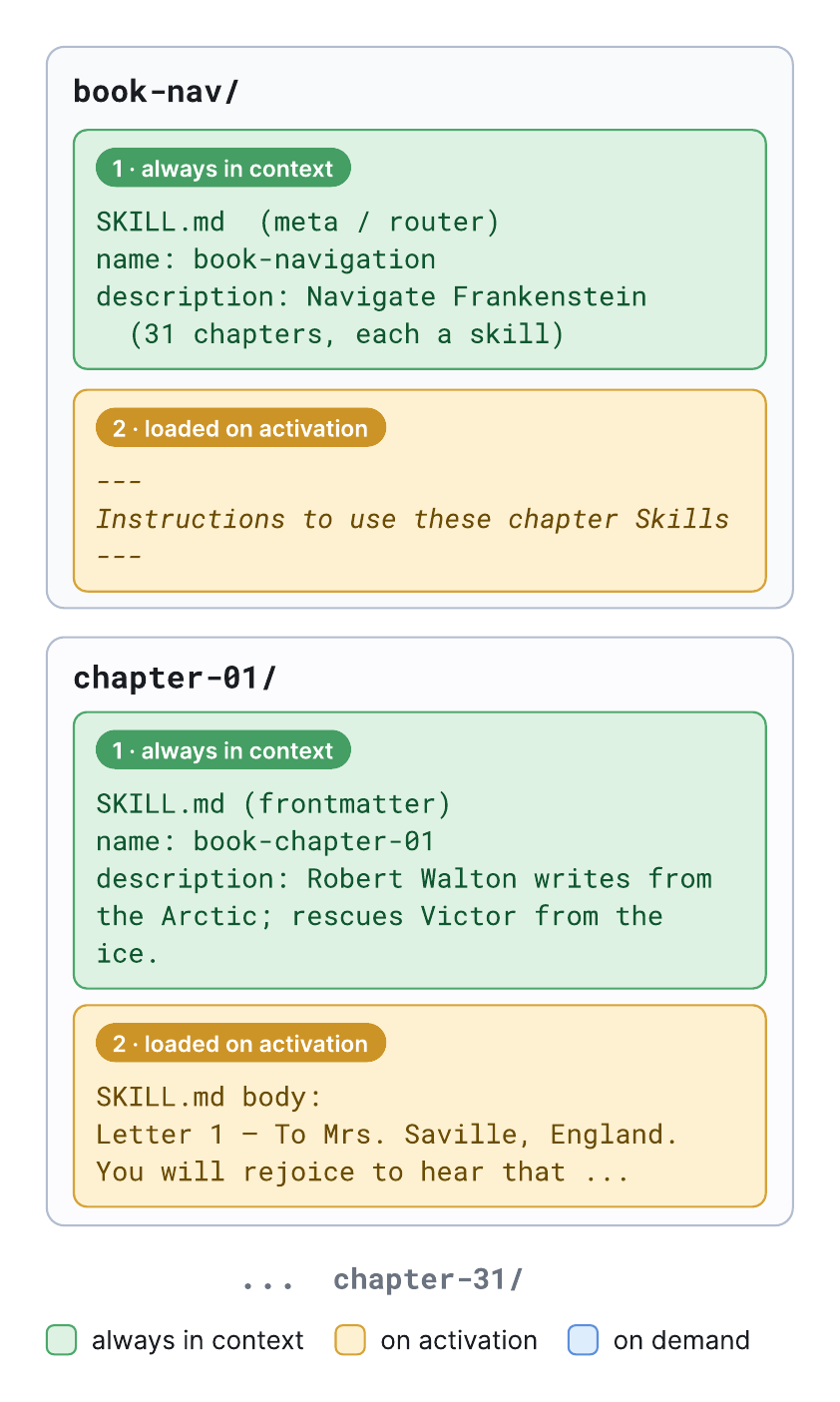}
    \caption{%
        The \hierdisc pack. %
    }
    \label{fig:hier}
\end{figure}

We state each approach precisely, including which artifacts stay in context versus load on demand, because the practitioner literature does not separate these.
The \flatdisc and \hierdisc packs (\autoref{fig:flat}, \autoref{fig:hier}) differ only in where each artifact sits on the disclosure timeline.

\paragraph{\rawdoc: raw-document navigation}
The first approach exposes no skill pack.
We place the raw book in the sandboxed file system beside the question and leave the agent's navigation policy unconstrained.
Depending on the harness, it may open the whole file, grep first, or read only matched passages.
This is the approach an agent falls back on when no structure is prepared for it, and the reference point against which we measure what disclosure adds.

\paragraph{\flatdisc: single-level disclosure}
The skill pack is a single skill rooted at one \skillmd{} file (\autoref{fig:flat}).
Its \texttt{description} field is always in context and describes the book as a whole, telling the agent to open the skill whenever a question concerns that title (\emph{discovery}).
Once the description matches the task, the agent reads the \skillmd{} body: a table that indexes every chunk by its path and its per-chunk description (\emph{activation}).
The agent then reads one or more chunk files from \texttt{references/} based on that table (\emph{execution}).
The per-chunk descriptions live in the body, so they pay context only after the book activates.

\paragraph{\hierdisc: recursive disclosure}
Each chunk becomes its own skill: a \skillmd{} whose \texttt{description} field is the chunk's description from \autoref{sec:method:pipeline} and whose body is the chunk content (\autoref{fig:hier}).
A meta-router at the top carries a book-level description and routes among the child skills.
The meta-router description and every child-skill description are always in context (\emph{discovery}).
The agent matches the task to one or more child descriptions and reads the corresponding child bodies on demand; for chunk content, \emph{activation} and \emph{execution} collapse, since the body is the content itself.
Reading a chunk is one step shorter than in \flatdisc, because every chunk is already a skill whose description sits in context.
\hierdisc differs from \flatdisc on two axes: the routing recurses one level deeper, and the per-chunk descriptions move from on-demand body context to always-loaded description context.

\subsection{What the comparison isolates}
\label{sec:method:isolation}

All three approaches hand the agent the identical task instruction over the same book (\appautoref{fig:task-prompts}) and score its written answer with the same deterministic verifier; only how the agent reaches a passage changes: raw navigation, flat disclosure, or hierarchical disclosure.
We measure the three approaches independently, then read two contrasts between their accuracies to answer the two questions from \autoref{sec:intro}.

The \rawdoc{} vs.\ \flatdisc{} contrast tests whether progressive disclosure helps at all on book-length understanding: whether a single description-gated load of indexed chunks beats giving the agent only the raw document and leaving navigation to it.
The \flatdisc{} vs.\ \hierdisc{} contrast tests whether \emph{depth} of disclosure matters: whether promoting per-chunk descriptions into always-loaded child descriptions and adding a meta-router improves routing enough to justify the always-loaded context tax.

%% file: src/environment.tex
\section{\loongdoc{}: a long-context agent environment}
\label{sec:env}

A static long-context reasoning benchmark scores one answer string and stops there.
It cannot watch the agent navigate, host a skill pack the agent must discover and open, or grow the corpus past a single book.
Our questions require all three.
We therefore rebuild \infbench{}~\citep{zhang2024infinitebench} into \loongdoc{}, an interactive environment.
Built on BenchFlow~\citep{benchflow2026}, \loongdoc{} runs any harness that implements the Agent Client Protocol (ACP)~\citep{acp2025} against a sandboxed task and records what it does.
Each \infbench{} question, multiple-choice (MC) or open QA, becomes a task the agent acts on through a file system rather than a fixed context, and every run leaves a scored trajectory we can inspect (\autoref{fig:environment}).
BenchFlow already hosts \textsc{SkillsBench}~\citep{li2026skillsbench}, a widely adopted benchmark for general-purpose agent skills~\citep{qwen2026qwen36,tencent2026hy3}; \loongdoc{} adds a long-context reasoning environment to the same runtime.

\input{Figures/environment.tex}

\paragraph{From a benchmark item to an agent task}
A \loongdoc{} task pairs a prompt with a sandboxed file system and a verifier that scores the result, the shape \textsc{Terminal-Bench}~\citep{merrill2026terminalbench} and \textsc{SkillsBench}~\citep{li2026skillsbench} establish for agent evaluation.
We adopt it to turn \infbench{} from a static probe into an agent environment, mapping one \infbench{} item onto one task.
We write the book into the sandbox as files, or the whole bundle at library scale, and hand the agent the question as its prompt.
Under the disclosure approaches of \autoref{sec:method}, we also stage a skill pack for the agent to discover through BenchFlow's native support for agent skills.
The verifier then reads the agent's final answer and returns a score.
Because the prompt says nothing about the book, the agent must reach the text through the file system, which makes raw-document access and progressive disclosure comparable conditions inside one harness.

\paragraph{Free-form navigation, recorded in full}
Because the book lives on disk, the agent chooses how to reach it, and any \mbox{ACP}-compatible harness can drive the environment.
In our runs, \codex, \piagent, and \claudecode read, grep, and open files at will rather than working from a context fixed in advance.
Every rollout returns the whole trajectory: the tool calls the agent issued, the files it opened, and the tokens it spent at each step.
This record lets us explain results, not just report them: the bare \codex trajectory, for instance, greps the raw text for the entities each question names and reads only the matched passages (\autoref{sec:results}), behavior the answer string alone would hide.

\paragraph{A score that doubles as a reward}
The verifier emits a scalar reward for every rollout, not only a corpus-level accuracy: closed-form questions score by normalized match and open questions by the \infbench{} open-QA metric, so each run carries its own per-trajectory reward.
Here we use that reward only to compare disclosure conditions.
But because the environment is sandboxed and the verifier reads only the agent's final answer, the same per-trajectory signal could post-train an agent, improving long-context reasoning with \loongdoc{} as the environment.

\paragraph{Navigation at library scale}
We push context length further by staging many books in one environment for the agent to search and read.
A single question still targets one book, so a larger bundle tests whether the agent can locate the answer-bearing book before reasoning over it.
A task stages $K$ books and asks about all of them.
Moving from $K{=}1$ to $K{=}20$ changes only how many files we write, not the interface the agent sees, and \autoref{sec:results:scaling} tests the agent's library-level navigation.

%% file: Figures/environment.tex
\begin{figure}[t]
\centering
\adjustbox{max width=\columnwidth}{%
\begin{tikzpicture}[
  >={Stealth[]},
  node distance=6.5mm,
  stage/.style={rounded corners=2pt, draw, thick, align=center,
                inner sep=4pt, minimum height=9mm, font=\small, text width=46mm},
  lbl/.style={font=\footnotesize},
]
  \node[stage, fill=blue!8]                   (item) {\infbench{} item: book(s) + question};
  \node[stage, fill=green!12, below=of item]  (fs)   {Sandbox file system\\[1pt]\footnotesize book files + book-to-skill pack};
  \node[stage, fill=orange!16, below=of fs]   (ag)   {Agent (\mbox{ACP} harness)\\[1pt]\footnotesize read / grep / open};
  \node[stage, fill=purple!12, below=of ag]   (vf)   {Verifier: answer match};
  \node[stage, fill=gray!18, below=of vf]     (rw)   {Reward + trajectory logged};

  \draw[->, thick] (item) -- node[lbl, right]{task} (fs);
  \draw[->, thick] ([xshift=-6mm]fs.south) -- node[lbl, left]{prompt} ([xshift=-6mm]ag.north);
  \draw[->, thick] ([xshift=6mm]ag.north) -- node[lbl, right]{files} ([xshift=6mm]fs.south);
  \draw[->, thick] (ag) -- node[lbl, right]{answer} (vf);
  \draw[->, thick] (vf) -- node[lbl, right]{score} (rw);
\end{tikzpicture}%
}
\caption{%
The \loongdoc{} environment: an \infbench{} book-QA item becomes a BenchFlow task the agent solves through a sandboxed file system.
}
\label{fig:environment}
\end{figure}
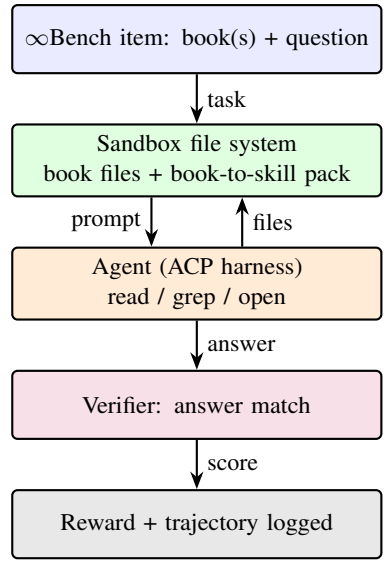

%% file: src/results.tex
\section{Results}
\label{sec:results}

\input{tables/infinitebench}

\subsection{Single-book question answering}
\label{sec:results:singlebook}
We evaluate the three approaches over single \infbench{}~\citep{zhang2024infinitebench} books across three harnesses, three models, and a \hybridrag baseline (\appautoref{sec:appendix:singlebook-setup}).
We report the results in \autoref{tab:infinitebench} and read two contrasts across its cells.

\paragraph{Disclosure helps where native navigation is weak}
The \rawdoc{} vs.\ \flatdisc{} contrast isolates the disclosure mechanism itself.
On \piagent and \claudecode, the \flatdisc skill pack matches or exceeds \rawdoc in every cell, the sole exception being \qwenmodel on \zhqa, where the two sit within one standard deviation, and the \haikumodel \enmc gain clears the high \rawdoc variance.
On these harnesses a single level of description-gated, in-skill-indexed chunk loading already beats letting the agent navigate the raw document on its own.

\paragraph{The gain is harness-dependent}
The \codex harness inverts this picture: on \gptmini the three approaches tie within error on all three subsets, so the skill pack that wins under \piagent adds nothing here.
Trajectory inspection explains the gap: under \rawdoc the bare \codex agent does not read the book linearly but builds its own retrieval, grepping the raw text for the entities named in each question and reading only the matched passages.
It reconstructs on the fly the locate-then-read capability that the skill pack pre-builds, so disclosure is redundant when the agent already navigates well natively; for such agents a pre-cut skill buys controllability of the retrieval path, not accuracy.

\paragraph{The gain is not reducible to retrieval}
The \hybridrag baseline tests whether the disclosure gain is just retrieval in disguise.
On \qwenmodel it trails both \rawdoc and \flatdisc across all three subsets, with the open-QA gap the widest, and on \gptmini it trails \flatdisc on \enmc as well.
Routing to a chunk through an in-skill index thus outscores the classical rerank-and-feed pipeline in nearly every cell.

\paragraph{Depth does not pay, and can hurt}
The \flatdisc{} vs.\ \hierdisc{} contrast isolates the value of pushing disclosure one level deeper, promoting the per-chunk index out of the activated parent body into always-loaded child-skill descriptions with a meta-router on top.
No cell rewards the extra depth: \hierdisc ties \flatdisc within noise on \haikumodel and on \codex \gptmini, and trails it everywhere else.
On \piagent it is actively harmful, collapsing \enmc from $0.9126$ under \flatdisc to $0.6398$ on \gptmini and falling below \flatdisc across all three subsets on \qwenmodel, steepest on \zhqa ($0.7479$ to $0.3890$); each drop lands far outside the cells' variance, as the always-loaded child descriptions saturate the router's context before it commits to a chunk.

\takeaway{On a single book, disclosure is a harness-dependent context-scaling tool, not a universal accuracy lever: it lifts weak-navigator harnesses (\piagent, \claudecode) and adds nothing to \codex, whose bare agent already reconstructs locate-then-read retrieval. Depth never helps on a single book and sometimes hurts, so one flat, in-skill-indexed level is the default; only at library scale, on weak-navigator open-QA cells, does the extra level recover (\autoref{sec:appendix:pi-scaling}).}

\subsection{Scaling from one book to a library}
\label{sec:results:scaling}
\input{tables/multibook}

\begin{figure}[t]
    \centering
    \includegraphics[width=\linewidth]{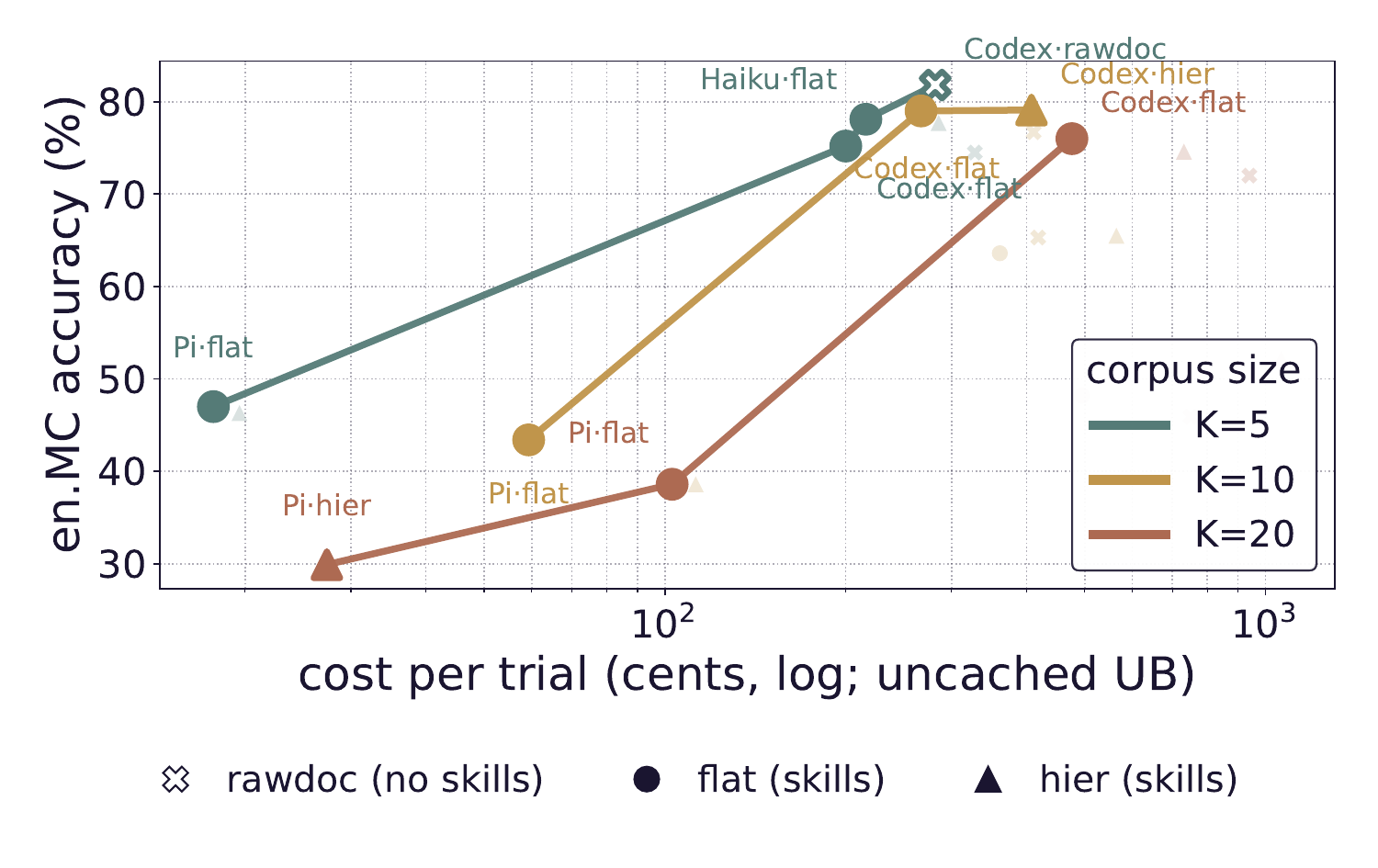}
    \caption{%
    Corpus-scaling cost--accuracy Pareto frontier on \enmc.}
    \label{fig:scaling-frontier}
\end{figure}
Single-book QA hands the agent the one document a question is about; a library-scale deployment instead asks it to first find the right book among many, then answer.
We probe this with the corpus-scale knob of \autoref{sec:env}: each multi-book bundle stages $K$ books and poses questions about all of them, each answerable from a single book, for $K \in \{5, 10, 20\}$ (\autoref{tab:multibook}).
We run \codex on \gptmini and \claudecode on \haikumodel across \enmc, \enqa, and \zhqa; the weaker \piagent harness supplies a further \enmc contrast (\autoref{sec:appendix:pi-scaling}).
We center \codex deliberately: it is the harness where disclosure bought nothing on single-book QA (\autoref{tab:infinitebench}), so if a skill pack earns its keep here, corpus scale breaks native navigation.
In \autoref{fig:scaling-frontier} we place these runs on a cost--accuracy plane (per-question cost from real token usage); we draw \enmc\ here and its \enqa\ companion, which carries the sharpest disclosure effect, in \autoref{fig:scaling-frontier-qa}, reading \zhqa\ from \autoref{tab:multibook}.
As $K$ grows the frontier recedes down and to the right: accuracy falls \emph{and} the cost of buying any fixed accuracy rises, so a larger corpus is strictly less efficient, not merely harder.
Its top point also shifts from \codex\,\rawdoc\ at $K{=}5$ to \codex\,\flatdisc\ at $K{\ge}10$, so the skill pack takes over the frontier only once the corpus is large; \codex\ holds the high-accuracy end, \piagent\ the low-cost end at roughly $10\times$ fewer tokens, and \claudecode\ is largely Pareto-dominated at $K{=}20$.

\paragraph{Corpus scale breaks the bare agent that single-book QA did not}
Even \codex loses accuracy as $K$ grows, and the loss is steepest on the open-QA subsets, where questions cannot be answered from parametric memory of a canonical novel.
Under \rawdoc, \enqa more than halves and \zhqa collapses to near zero from $K{=}5$ to $K{=}20$: the same bare agent that matched the skill pack on a single book cannot reliably find the target book once the corpus grows to twenty.
\enmc degrades more gently, reflecting the pre-training confound below.

\paragraph{Disclosure rescues scaling where it was redundant single-book}
The \flatdisc skill pack degrades more slowly than \rawdoc, and the two diverge as $K$ grows.
At $K{=}5$, where the whole corpus still fits comfortably in context, \flatdisc carries a small routing overhead and trails \rawdoc on \enmc; by $K{=}20$ it leads on all three subsets.
The \enqa $K{=}20$ gap clears one standard error ($0.462$ vs.\ $0.257$) and the \enmc gap is borderline; only \zhqa stays directional at this sample size.
The mechanism is the corpus-routing analogue of the single-book story: a pre-cut skill index lets the agent commit to the target book before it burns context traversing the library, exactly the failure mode that sinks \rawdoc at large $K$.

\paragraph{The rescue depends on routing depth, not on disclosure alone}
The rescue is specific to the \flatdisc pack; \hierdisc does not reproduce it.
On \enmc the two disclosure packs track each other and both clear \rawdoc at $K{=}20$, but on open QA they part: at $K{=}20$ on \enqa, \hierdisc collapses back onto \rawdoc while \flatdisc holds well over $1.7\times$ their accuracy.
The mechanism follows from what each pack keeps in context: \hierdisc always loads every chunk's description, so a twenty-book library inflates the always-on budget and recreates the context pressure that disclosure is meant to relieve, whereas \flatdisc pays for a per-chunk summary only when it reads that chunk.
At library scale the flat single-index pack is therefore the better default, and the single-book verdict that the two are interchangeable does not carry over.

\paragraph{A second harness reproduces the \enqa rescue on a weaker model}
The full \claudecode grid on \haikumodel, a weaker model than \gptmini, reproduces the \enqa pro-skill signal at every scale: \flatdisc leads \rawdoc across all three $K$, with the $K{=}10$ gap clearing two standard errors, so the rescue survives a change of both harness and model.
\hierdisc again fails to reproduce it, tracking \rawdoc at every $K$ while \flatdisc stays ahead.
The other two subsets cut the other way. On \enmc the three configurations stay within a standard error at every $K$ on both grids, the memorized-novel confound blunting any disclosure signal; but on the weaker \piagent harness \flatdisc leads \rawdoc on \enmc at every $K$ (\autoref{tab:multibook-pi}), further evidence that the flat pack is the best default.
On \zhqa the skill is neutral-to-negative, and at $K{=}20$ \rawdoc leads; this cell stays directional at our sample size, and the multi-book gap tracks base-model capability rather than the skill pack.

\takeaway{At library scale the flat skill pack becomes an accuracy tool even for the strong \codex agent: it beats the bare agent in all six \enqa cells across two harnesses and two models, the most robust effect in the table. The payoff is task- and language-specific, sharpest on English open QA and absent on Chinese, where the base model, not the skill pack, sets the ceiling.}

%% file: tables/infinitebench.tex
\begin{table*}[t]
\centering
\small
\caption{Single-book \infbench{} accuracy across agent harnesses, models, and skill-pack configurations.
Cells give mean accuracy $\pm$ standard deviation across three seeds.
\textbf{Bold} marks the best method within each (harness, model) block; when the top two methods differ by less than a standard deviation we bold both as a tie. \underline{Underline} marks the best score in each task column across all settings.
}
\label{tab:infinitebench}
\adjustbox{max width=\textwidth}{%
\begin{tabular}{lllccc}
\toprule
Harness & Model & Method & \enmc & \enqa & \zhqa \\
\midrule
\multirow{3}{*}{\codex}
  & \multirow{3}{*}{\gptmini}
    & \cellcolor{rowraw}\rawdoc  & \cellcolor{rowraw}$0.8943 \pm 0.0174$ & \cellcolor{rowraw}$0.7412 \pm 0.0219$ & \cellcolor{rowraw}$\underline{\mathbf{0.8652}} \pm 0.0220$ \\
  & & \cellcolor{rowflat}\flatdisc & \cellcolor{rowflat}$\mathbf{0.8977} \pm 0.0177$ & \cellcolor{rowflat}$\underline{\mathbf{0.7516}} \pm 0.0210$ & \cellcolor{rowflat}$0.8390 \pm 0.0238$ \\
  & & \cellcolor{rowhier}\hierdisc & \cellcolor{rowhier}$0.8874 \pm 0.0139$ & \cellcolor{rowhier}$0.7377 \pm 0.0242$ & \cellcolor{rowhier}$0.8525 \pm 0.0237$ \\
\midrule
\multirow{8}{*}{\piagent}
  & \multirow{4}{*}{\gptmini}
    & \cellcolor{rowraw}\rawdoc  & \cellcolor{rowraw}$0.8851 \pm 0.0053$ & \cellcolor{rowraw}$0.7161 \pm 0.0329$ & \cellcolor{rowraw}$0.6856 \pm 0.0163$ \\
  & & \cellcolor{rowflat}\flatdisc & \cellcolor{rowflat}$\underline{\mathbf{0.9126}} \pm 0.0208$ & \cellcolor{rowflat}$\mathbf{0.7259} \pm 0.0062$ & \cellcolor{rowflat}$\mathbf{0.7007} \pm 0.0331$ \\
  & & \cellcolor{rowhier}\hierdisc & \cellcolor{rowhier}$0.6398 \pm 0.0043$ & \cellcolor{rowhier}$0.5120 \pm 0.0060$ & \cellcolor{rowhier}$0.6510 \pm 0.0390$ \\
  & & \cellcolor{rowrag}\hybridrag & \cellcolor{rowrag}$0.7628 \pm 0.0066$ & \cellcolor{rowrag}$0.6214 \pm 0.0091$ & \cellcolor{rowrag}$\mathbf{0.7020} \pm 0.0136$ \\
\cmidrule(lr){2-6}
  & \multirow{4}{*}{\qwenmodel}
    & \cellcolor{rowraw}\rawdoc  & \cellcolor{rowraw}$0.7865 \pm 0.0355$ & \cellcolor{rowraw}$0.6813 \pm 0.0644$ & \cellcolor{rowraw}$\mathbf{0.7563} \pm 0.0177$ \\
  & & \cellcolor{rowflat}\flatdisc & \cellcolor{rowflat}$\mathbf{0.8023} \pm 0.0799$ & \cellcolor{rowflat}$\mathbf{0.6913} \pm 0.0180$ & \cellcolor{rowflat}$0.7479 \pm 0.0287$ \\
  & & \cellcolor{rowhier}\hierdisc & \cellcolor{rowhier}$0.6460 \pm 0.0696$ & \cellcolor{rowhier}$0.5933 \pm 0.0040$ & \cellcolor{rowhier}$0.3890 \pm 0.1388$ \\
  & & \cellcolor{rowrag}\hybridrag & \cellcolor{rowrag}$0.7470 \pm 0.0086$ & \cellcolor{rowrag}$0.5439 \pm 0.0000$ & \cellcolor{rowrag}$0.5856 \pm 0.0038$ \\
\midrule
\multirow{3}{*}{\claudecode}
  & \multirow{3}{*}{\haikumodel}
    & \cellcolor{rowraw}\rawdoc  & \cellcolor{rowraw}$0.7448 \pm 0.1166$ & \cellcolor{rowraw}$0.6892 \pm 0.0305$ & \cellcolor{rowraw}$0.8142 \pm 0.0077$ \\
  & & \cellcolor{rowflat}\flatdisc & \cellcolor{rowflat}$0.8667 \pm 0.0170$ & \cellcolor{rowflat}$0.7173 \pm 0.0086$ & \cellcolor{rowflat}$\mathbf{0.8330} \pm 0.0062$ \\
  & & \cellcolor{rowhier}\hierdisc & \cellcolor{rowhier}$\mathbf{0.8687} \pm 0.0267$ & \cellcolor{rowhier}$\mathbf{0.7255} \pm 0.0101$ & \cellcolor{rowhier}$0.8295 \pm 0.0133$ \\
\bottomrule
\end{tabular}%
}
\end{table*}

%% file: tables/multibook.tex
\begin{table*}[t]
\centering
\small
\caption{Multi-book corpus-scaling accuracy on \infbench{}. Conventions follow \autoref{tab:infinitebench}, except that cells give mean $\pm$ standard error across bundles and seeds rather than seed-level standard deviation.
}
\label{tab:multibook}
\begin{tabular}{lllccc}
\toprule
Harness & Task & Method & $K{=}5$ & $K{=}10$ & $K{=}20$ \\
\midrule
\multirow{9}{*}{\shortstack[l]{\codex\\\gptmini}}
  & \multirow{3}{*}{\enmc}
    & \cellcolor{rowraw}\rawdoc  & \cellcolor{rowraw}$\mathbf{0.818} \pm 0.019$ & \cellcolor{rowraw}$0.767 \pm 0.027$ & \cellcolor{rowraw}$0.720 \pm 0.019$ \\
  & & \cellcolor{rowflat}\flatdisc & \cellcolor{rowflat}$0.752 \pm 0.034$ & \cellcolor{rowflat}$0.790 \pm 0.021$ & \cellcolor{rowflat}$\mathbf{0.760} \pm 0.014$ \\
  & & \cellcolor{rowhier}\hierdisc & \cellcolor{rowhier}$0.777 \pm 0.043$ & \cellcolor{rowhier}$\mathbf{0.791} \pm 0.018$ & \cellcolor{rowhier}$0.746 \pm 0.017$ \\
\cmidrule(lr){2-6}
  & \multirow{3}{*}{\enqa}
    & \cellcolor{rowraw}\rawdoc  & \cellcolor{rowraw}$0.657 \pm 0.041$ & \cellcolor{rowraw}$0.577 \pm 0.044$ & \cellcolor{rowraw}$0.257 \pm 0.067$ \\
  & & \cellcolor{rowflat}\flatdisc & \cellcolor{rowflat}$\mathbf{0.708} \pm 0.023$ & \cellcolor{rowflat}$0.582 \pm 0.045$ & \cellcolor{rowflat}$\mathbf{0.462} \pm 0.061$ \\
  & & \cellcolor{rowhier}\hierdisc & \cellcolor{rowhier}$0.667 \pm 0.030$ & \cellcolor{rowhier}$\mathbf{0.606} \pm 0.018$ & \cellcolor{rowhier}$0.267 \pm 0.069$ \\
\cmidrule(lr){2-6}
  & \multirow{3}{*}{\zhqa}
    & \cellcolor{rowraw}\rawdoc  & \cellcolor{rowraw}$0.698 \pm 0.074$ & \cellcolor{rowraw}$\mathbf{0.214} \pm 0.092$ & \cellcolor{rowraw}$0.043 \pm 0.022$ \\
  & & \cellcolor{rowflat}\flatdisc & \cellcolor{rowflat}$0.738 \pm 0.056$ & \cellcolor{rowflat}$0.178 \pm 0.079$ & \cellcolor{rowflat}$0.095 \pm 0.043$ \\
  & & \cellcolor{rowhier}\hierdisc & \cellcolor{rowhier}$\mathbf{0.771} \pm 0.046$ & \cellcolor{rowhier}$0.137 \pm 0.070$ & \cellcolor{rowhier}$\mathbf{0.097} \pm 0.043$ \\
\midrule
\multirow{9}{*}{\shortstack[l]{\claudecode\\\haikumodel}}
  & \multirow{3}{*}{\enmc}
    & \cellcolor{rowraw}\rawdoc  & \cellcolor{rowraw}$0.745 \pm 0.024$ & \cellcolor{rowraw}$0.653 \pm 0.016$ & \cellcolor{rowraw}$0.459 \pm 0.047$ \\
  & & \cellcolor{rowflat}\flatdisc & \cellcolor{rowflat}$\mathbf{0.781} \pm 0.033$ & \cellcolor{rowflat}$0.636 \pm 0.035$ & \cellcolor{rowflat}$0.482 \pm 0.037$ \\
  & & \cellcolor{rowhier}\hierdisc & \cellcolor{rowhier}$0.777 \pm 0.021$ & \cellcolor{rowhier}$\mathbf{0.655} \pm 0.020$ & \cellcolor{rowhier}$\mathbf{0.521} \pm 0.043$ \\
\cmidrule(lr){2-6}
  & \multirow{3}{*}{\enqa}
    & \cellcolor{rowraw}\rawdoc  & \cellcolor{rowraw}$0.582 \pm 0.038$ & \cellcolor{rowraw}$0.413 \pm 0.021$ & \cellcolor{rowraw}$0.301 \pm 0.037$ \\
  & & \cellcolor{rowflat}\flatdisc & \cellcolor{rowflat}$\mathbf{0.642} \pm 0.023$ & \cellcolor{rowflat}$\mathbf{0.480} \pm 0.025$ & \cellcolor{rowflat}$\mathbf{0.354} \pm 0.045$ \\
  & & \cellcolor{rowhier}\hierdisc & \cellcolor{rowhier}$0.589 \pm 0.027$ & \cellcolor{rowhier}$0.453 \pm 0.032$ & \cellcolor{rowhier}$0.302 \pm 0.035$ \\
\cmidrule(lr){2-6}
  & \multirow{3}{*}{\zhqa}
    & \cellcolor{rowraw}\rawdoc  & \cellcolor{rowraw}$\mathbf{0.730} \pm 0.025$ & \cellcolor{rowraw}$0.550 \pm 0.027$ & \cellcolor{rowraw}$\mathbf{0.499} \pm 0.069$ \\
  & & \cellcolor{rowflat}\flatdisc & \cellcolor{rowflat}$0.696 \pm 0.030$ & \cellcolor{rowflat}$\mathbf{0.576} \pm 0.023$ & \cellcolor{rowflat}$0.401 \pm 0.046$ \\
  & & \cellcolor{rowhier}\hierdisc & \cellcolor{rowhier}$0.687 \pm 0.029$ & \cellcolor{rowhier}$0.533 \pm 0.040$ & \cellcolor{rowhier}$0.389 \pm 0.045$ \\
\bottomrule
\end{tabular}
\end{table*}

%% file: src/conclusion.tex
\section{Conclusion}
\label{sec:conclusion}

We ran the first controlled study of progressive disclosure for agentic long-document reasoning,
comparing raw-document navigation, flat and hierarchical skill packs, and a hybrid retriever on \infbench{} across three harnesses and three model families.
Progressive disclosure buys context, not intelligence: on a single book a strong agent already greps the passages it needs, so disclosure adds nothing, but once the corpus outgrows what native navigation can hold, flat disclosure becomes decisive.
A second routing level never reproduces this gain and sometimes breaks accuracy.
For packaging book-length material the guidance is concrete: package a book as one skill layered progressive disclosure,
not as a multiple parallel packages of child skills with always-loaded description.

%% file: src/limitations.tex
\section*{Limitations}
\label{sec:limitations}

Our study leaves several threats open.

\paragraph{Pre-training confound on \enmc}
The clearest threat is the pre-training confound on \enmc: the \infbench{} multiple-choice books are canonical English novels the models have likely memorized, so the bare agent often answers from parametric knowledge rather than from reading, which blunts any disclosure signal on that subset.
Two traces sharpen the worry: even at $K{=}20$ the \rawdoc agent answers \enmc correctly in a handful of tool calls, far too few to have read twenty books, and the single-book \codex trajectories show the agent recognizing a renamed book as a canonical work.
The open-QA subsets, where memory helps far less, are the cleaner test; a held-out or synthetic-book corpus would remove the confound.

\paragraph{Narrow scaling axis}
We sample the corpus size only at $K \in \{1, 5, 10, 20\}$ within a single benchmark family, so we bracket the point where disclosure begins to pay off only coarsely, and we cannot say whether the effect transfers to non-narrative corpora such as code or technical manuals.

\paragraph{Task- and language-specific gains}
The gains are task- and language-specific rather than universal, sharpest on English open QA and absent or negative on Chinese, where the underlying model, not the skill pack, sets the ceiling: \haikumodel far outscores \gptmini on \zhqa at $K{=}20$, the largest raw-capability gap in the multi-book table.

\paragraph{Fixed recipe}
We hold the chunking and description recipe fixed at one widely adopted book-to-skill procedure, so a different chunk granularity or description-writing model might shift the balance between the three approaches.

\paragraph{Sample size}
Our sample sizes are those \infbench{} provides, so several of our comparisons stay within a standard error.
This bound comes from the dataset, not from \loongdoc: the pipeline converts any long-context reading benchmark into the same agentic file-system environment, so a benchmark with more questions would tighten every estimate.
We build on \infbench{} because it ships gold answers that a local deterministic verifier scores directly, whereas comparable benchmarks such as NovelQA~\citep{wang2024novelqa} withhold their labels and score predictions only through a third-party platform (Codabench), which we could not fit into a controlled multi-harness sweep.

%% file: src/ethics.tex
\section*{Ethics Statement}
\label{sec:ethics}

Our study raises no direct ethical concerns.
We evaluate reading strategies on \infbench{}~\citep{zhang2024infinitebench}, a public benchmark of long documents used under its released terms, and we introduce no new human-subject data.
The books behind every subset, multiple-choice and open QA alike, are copyrighted novels.
We neither redistribute their text nor release the derived skill packs; \loongdoc{} builds each task and pack from the public \infbench{} release, so reproducing our results adds no copyrighted material beyond what the benchmark already distributes.

%% file: src/appendix.tex
\appendix

\section{Methodology details}

\subsection{Task instructions and scoring}
\label{sec:appendix:taskprompts}

Every approach the paper compares runs the same task instruction (\autoref{fig:task-prompts}), which tells the agent to read each question and write its answer to a per-question file, and states where to find the book.
A book-QA item takes one of two instructions: multiple-choice for \enmc and free-form for \enqa and \zhqa.
Holding the instruction fixed across \rawdoc, \flatdisc, and \hierdisc leaves the file system beside the text as the only variable, so a score difference reflects how the agent reached a passage rather than how it was asked (\autoref{sec:method:isolation}).

A deterministic verifier then scores the written answers, matching the official \infbench{} \texttt{compute\_scores.py}: it extracts the option letter for multiple choice by regular expression and compares it to the gold label, and it scores free-form answers by \infbench{}'s normalized match and token-level F1, with the Chinese variant normalizing Chinese punctuation and whitespace.
No answer-extraction model sits between the agent and the score, so the reward depends only on the agent's answer file, not on a second model's reading of it.

\input{Figures/task-prompts}

\paragraph{Constructing a library-scale task}
A multi-book task reuses the single-book instruction, prepends one framing block, and relabels the file system (\autoref{fig:sandbox-layout}).
The framing block states that the corpus holds $K$ labeled books, that different questions concern different books, and that the agent must identify which book a question asks about before answering; a \texttt{corpus-index.md} gives a one-line summary of each member book.
Every question and answer identifier gains the book label as a prefix, so the verifier keys each gold answer on the full prefixed identifier and the deterministic scoring of the single-book case carries over unchanged.

\input{Figures/sandbox-layout}

\subsection{Description-generation prompts}
\label{sec:appendix:prompts}

The book-to-skill pipeline (\autoref{sec:method:pipeline}) generates all routing metadata with the two prompts in \autoref{fig:prompts}.
The per-chunk prompt runs once per chunk; the book-level prompt runs once per book over the resulting chunk descriptions.
Both descriptions are fixed inputs to the two disclosure approaches, so the same text routes the agent in \flatdisc and \hierdisc alike.
The choices below fix the metadata but do not bear on the disclosure question the paper studies; we record them for reproducibility.

\input{Figures/prompts}

\paragraph{Two-part per-chunk output}
The per-chunk prompt asks for a one- or two-sentence summary and a comma-separated list of key elements, the named characters, locations, organizations, and topics the chunk introduces.
We parse the two fields from the \texttt{SUMMARY:} and \texttt{KEY\_ELEMENTS:} markers; when a response carries neither marker, we keep the whole response as the summary and leave the key-element list empty.
The two fields become adjacent columns of the \flatdisc table of contents and, concatenated, the \texttt{description} of each \hierdisc chapter skill.
The key-element list gives the router exact surface forms, such as proper names and place names, that a one-sentence summary tends to paraphrase away.

\paragraph{Book-level summary}
The book-level prompt reads only the chapter summaries, never the book text, and writes the one- or two-sentence description that gates discovery.
It works from anonymized names and must not name or infer the book's title or author, so a title the agent happens to recognize cannot shortcut retrieval and inflate a score.

\paragraph{Determinism}
Every call runs at temperature \num{0} with a fixed seed, so rerunning the pipeline on the same book reproduces the same metadata.
A per-book manifest records the model, seed, and a hash of each prompt, letting a later run reuse an existing skill pack rather than regenerate it.

\subsection{Hybrid RAG for agents}

Hybrid RAG is our conventional retrieval-indexing baseline for long-document agent evaluation. We describe here how it builds the book index, fuses sparse and dense retrieval, and feeds the retrieved evidence to the answer model (\autoref{fig:hybrid-rag-pipeline}), separating what it computes while indexing from what it computes at query time.
\input{Figures/hybrid-rag-pipeline}

\paragraph{Index construction}
We split a book into passages, give each a stable chunk identifier, and build two retrieval views over the chunk set. A BM25 index forms the sparse view, preserving exact lexical matches for names, quoted phrases, and local events; the BGE-M3 embedding model encodes each chunk into the dense view. We store each index under a content hash of the source book with a manifest recording the embedding model, reranker model, and build metadata, so repeated trials over the same book load the existing index instead of rebuilding it.

\paragraph{Hybrid retrieval}
At query time, the two retrievers run in parallel over the chunk set: BM25 scores rank chunks for the sparse retriever, and embedding similarity to the question ranks them for the dense retriever. Reciprocal rank fusion (RRF) merges the two ranked lists, a BGE cross-encoder reranks the fused candidates, and the top chunks enter the answer prompt as evidence. We fix the candidate-retrieval and reranking budgets across runs.

\paragraph{Agent interface}
The agent never sees the full book. For each question it receives a prompt carrying the retrieved evidence chunks and the benchmark question, and we instruct it to output exactly one option letter for multiple choice and only the final answer text for free-form QA, including Chinese QA. We save the retrieved and reranked chunk identifiers for every query, so we can inspect the evidence the agent used without relying on hidden model state.

\paragraph{Reuse and accounting}
Book content, not the answer model, keys the retrieval index, so the same index serves repeated trials and different answer models without leaking answers. We report index construction separately from per-query answer generation: the one-time indexing cost amortizes over every question for a book, while we measure answer-generation latency and token cost at query time.

\section{Experimental setup}

\paragraph{Compute environment}
The closed models (\gptmini, \haikumodel, and the \codex\ backend) answer as hosted API calls, so the harnesses only orchestrate requests; we run that orchestration in local Docker containers at their default resource classes.
Only \qwenmodel\ needs local hardware: we serve it with vLLM on two NVIDIA RTX 8000 GPUs, sharded with tensor parallelism (\texttt{--tensor-parallel-size~2}) and a \num{131072}-token context window.
Across every adapter we pin the agent packages to \texttt{pi-coding-agent}~0.66.1 and \texttt{pi-acp}~0.0.25, so agent-behavior changes cannot confound the cross-harness comparison.

\subsection{Single-book evaluation}
\label{sec:appendix:singlebook-setup}

We evaluate the three approaches from \autoref{sec:method:approaches} on three \infbench{}~\citep{zhang2024infinitebench} subsets that exercise book-length understanding: English multiple-choice (\enmc), English open QA (\enqa), and Chinese open QA (\zhqa).
Three agent harnesses (\codex, \piagent, and \claudecode) drive three models: \gptmini, \qwenmodel, and \haikumodel.
Each harness runs \rawdoc, \flatdisc, and \hierdisc over the same book, so the only variable is how the agent reaches a passage (\autoref{sec:method:isolation}).
All runs set \concurrenttask to avoid Docker timeouts.

\paragraph{Retrieval baseline}
The three agentic approaches all keep the document in the agent's reach and let it read on demand.
To place them against the classical alternative, we add \hybridrag, which replaces in-context reading with retrieval-augmented generation: it builds a per-document index, reranks the passages most relevant to the question, and feeds only that evidence to the answer model.
We collect \hybridrag on both \qwenmodel and \gptmini across all three subsets as a second-model check.

\section{Additional experimental results}

\subsection{Corpus scaling on the \piagent harness}
\label{sec:appendix:pi-scaling}
\input{tables/multibook-pi}
In \autoref{tab:multibook-pi} we add the \piagent harness to the corpus-scaling study of \autoref{sec:results:scaling}, run on \gptmini in the same setting as \autoref{tab:multibook}.
For \piagent we report all three subsets; \enmc is the axis it shares with the \codex grid of \autoref{tab:multibook}, so the strong- and weak-navigator harnesses meet there.
\codex is the strong-navigator end of the scaling story and \piagent the weak-navigator end, and the two bracket the harness-dependence claim.
On \enmc, \flatdisc leads \rawdoc at every $K$ under \piagent, whereas under \codex the same pack trails at $K{=}5$ and only pulls ahead once the corpus is large.
The weaker the agent's native navigation, the earlier the skill pack earns its keep.
On \enmc, \piagent also echoes the routing-depth result: its \hierdisc pack is the worst configuration at $K{=}20$, below even \rawdoc, under the same always-loaded-index pressure that sinks \hierdisc on \codex open QA.
Open QA reverses this depth ordering.
On \zhqa, \flatdisc drags below \rawdoc at every $K$ and collapses to $0.137$ at $K{=}20$, while \hierdisc holds the best cell there at $0.330$; on \enqa, \hierdisc leads at $K{=}10$ and stays above \rawdoc at $K{=}20$.
Where the corpus is large and the questions are open, the extra routing level that hurts \enmc recovers, so depth is a scale- and task-specific effect rather than a uniform cost.

\subsection{The \enqa\ scaling frontier}
\label{sec:appendix:qa-frontier}
\input{Figures/scaling-frontier-qa}
The main text draws \autoref{fig:scaling-frontier} for \enmc\ and reads the disclosure effect on \enqa\ from \autoref{tab:multibook}; \autoref{fig:scaling-frontier-qa} redraws that plane for \enqa, where the effect is sharpest, so the frontier can carry the claim visually.
The rescue that the table reports as numbers becomes geometry: at $K{=}20$ the flat pack alone holds the high-accuracy end of the frontier ($0.462$), while \rawdoc and \hierdisc collapse together to $0.257$ and $0.267$, twice as far down the accuracy axis.
The gap is wider than anything on the \enmc\ plane, where the three configurations stay within a standard error at every $K$ (\autoref{sec:appendix:enmc-scale}); \enqa\ is the subset on which disclosure separates from bare navigation.
Neither harness is scaling-robust here the way \codex is on \enmc: every curve recedes down and to the right with $K$, because open QA cannot fall back on parametric memory of a canonical novel, yet the ordering at each $K$ still favors \flatdisc.
Cost sharpens the verdict. At $K{=}20$ \rawdoc reads the most tokens ($68.3$M/question, an uncached upper bound of $\$52$) for the worst accuracy, the bare agent burning context re-reading the library and still misrouting.
\flatdisc reaches nearly double the accuracy at roughly half the tokens and half the cost ($32.5$M, $\$25$): at library scale the flat pack is both the more accurate and the cheaper choice.

\subsection{\enmc under the memorized-novel confound at scale}
\label{sec:appendix:enmc-scale}
On \enmc the three configurations stay within a standard error at every $K$ under both the \codex and \claudecode grids, so no approach separates cleanly.
The memorized-novel confound (see \hyperref[sec:limitations]{Limitations}) blunts the comparison at the smaller scales; only at $K{=}20$ does \hierdisc edge ahead in point estimate, a weak hint that broad traversal of the always-loaded index helps once the corpus is largest.
We read this as directional, not as a claim.

\subsection{Cost accounting for the scaling frontier}
\label{sec:appendix:cost}
The per-question cost on the cost--accuracy plane of \autoref{fig:scaling-frontier} and \autoref{fig:scaling-frontier-qa} is an uncached upper bound: it sums real per-call token usage with no prompt caching.
The always-loaded descriptions that \flatdisc and \hierdisc carry are the most cache-friendly part of the input, repeated across every question about a book, so caching them would cut disclosure's cost more than \rawdoc's and push disclosure further ahead on the frontier.
We report the uncached figure as the conservative case.

%% file: Figures/task-prompts.tex
\begin{figure*}[t]
    \centering
    \begin{tcolorbox}[
        colback=ForestGreen!5,
        colframe=ForestGreen!60,
        title=\textbf{Multiple-choice task instruction (\enmc)},
        fonttitle=\normalfont,
        boxrule=0.5pt,
        left=4pt, right=4pt, top=3pt, bottom=3pt,
    ]
    \footnotesize\ttfamily
    \begin{verbatim}
You are given the full text of a book and a set of
multiple-choice questions about it.

The full book text is available at /environment/book.txt.

Answer every question by choosing one of the provided options.
Each question is stored as a separate file under
/environment/questions/; each subdirectory (e.g. 42/) contains
a question.md with the question text and its answer options.
List the directory to discover all question IDs.

Output format: for each question, write your chosen option
letter (A, B, C, or D) to a plain-text file at
/logs/artifacts/questions/<QID>/answer.txt. Write one
answer.txt for every question.
\end{verbatim}
    \end{tcolorbox}

    \begin{tcolorbox}[
        colback=ForestGreen!5,
        colframe=ForestGreen!60,
        title=\textbf{Free-form QA task instruction (\enqa, \zhqa)},
        fonttitle=\normalfont,
        boxrule=0.5pt,
        left=4pt, right=4pt, top=3pt, bottom=3pt,
    ]
    \footnotesize\ttfamily
    \begin{verbatim}
You are given the full text of a book and a set of questions
about it.

The full book text is available at /environment/book.txt.

Answer every question with a short, accurate answer (one or a
few words). Each question is stored as a separate file under
/environment/questions/; each subdirectory (e.g. 42/) contains
a question.md with the question text. List the directory to
discover all question IDs.

Output format: for each question, write your answer as plain
text to /logs/artifacts/questions/<QID>/answer.txt. Write one
answer.txt for every question.
\end{verbatim}
    \end{tcolorbox}
    \caption{%
        The task instructions handed to the agent, held fixed across the three approaches (\autoref{sec:method:isolation}).
        A book-QA item maps to one of two instructions: multiple-choice for \enmc and free-form for \enqa and \zhqa (both use the same English instruction).
        The instruction carries no placeholders; only the artifacts beside \texttt{book.txt} in the file system change across \rawdoc, \flatdisc, and \hierdisc.
        At library scale the same instruction gains a short bundle preamble and rewrites the book path to \texttt{/environment/books/<label>/book.txt}.
        The verifier scores the written answers deterministically, so no answer-extraction model sits between the agent and the score.
    }
    \label{fig:task-prompts}
\end{figure*}

%% file: Figures/sandbox-layout.tex
\begin{figure*}[t]
    \centering
    \begin{tcolorbox}[
        colback=RoyalBlue!4,
        colframe=RoyalBlue!55,
        title=\textbf{Single-book task},
        fonttitle=\normalfont,
        boxrule=0.5pt,
        left=4pt, right=4pt, top=3pt, bottom=3pt,
    ]
    \footnotesize\ttfamily
    \begin{verbatim}
/environment/                 read-only mount, staged by the adapter
|-- book.txt                  full book text
`-- questions/
    |-- 42/question.md        question text (plus options for MC)
    `-- .../                  one directory per question ID

/logs/artifacts/questions/    written by the agent
`-- 42/answer.txt             one answer.txt per question ID
\end{verbatim}
    \end{tcolorbox}

    \begin{tcolorbox}[
        colback=RoyalBlue!4,
        colframe=RoyalBlue!55,
        title=\textbf{Library-scale ($K$-book) task},
        fonttitle=\normalfont,
        boxrule=0.5pt,
        left=4pt, right=4pt, top=3pt, bottom=3pt,
    ]
    \footnotesize\ttfamily
    \begin{verbatim}
/environment/                 read-only mount, staged by the adapter
|-- corpus-index.md           one-line summary of each member book
|-- books/
|   |-- A/book.txt            full text of book A
|   `-- B/book.txt            full text of book B
`-- questions/
    |-- A_42/question.md      question IDs carry the book label
    `-- B_157/question.md

/logs/artifacts/questions/    written by the agent
|-- A_42/answer.txt           answers carry the same label prefix
`-- B_157/answer.txt
\end{verbatim}
    \end{tcolorbox}
    \caption{%
        Sandbox layout the agent sees, for a single-book task (top) and a $K$-book library task (bottom).
        The adapter stages the read-only \texttt{/environment} tree; the agent writes one \texttt{answer.txt} per question under \texttt{/logs/artifacts/questions}, and the deterministic verifier reads those files (\autoref{sec:appendix:taskprompts}).
        Only the file system changes across \rawdoc, \flatdisc, and \hierdisc; the routing artifacts sit beside \texttt{book.txt} and are omitted here.
        At library scale the books move under \texttt{/environment/books/<label>/}, a \texttt{corpus-index.md} lists them, and every question and answer identifier gains the book label as a prefix.
    }
    \label{fig:sandbox-layout}
\end{figure*}

%% file: Figures/prompts.tex
\begin{figure*}[t]
    \centering
    \begin{tcolorbox}[
        colback=RoyalBlue!5,
        colframe=RoyalBlue!60,
        title=\textbf{Per-chunk description prompt},
        fonttitle=\normalfont,
        boxrule=0.5pt,
        left=4pt, right=4pt, top=3pt, bottom=3pt,
    ]
    \footnotesize\ttfamily
    \begin{verbatim}
You are summarizing one section of a long book to help an
agent navigate the book later. Read the section below and
produce:

1. A one- or two-sentence summary of what this section covers.
2. A comma-separated list of key elements: named characters,
   locations, organizations, or central topics that appear in
   this section.

Output format (exactly this; no extra prose, no Markdown
headers):

SUMMARY: <one to two sentences>
KEY_ELEMENTS: <comma-separated list>
\end{verbatim}
    \end{tcolorbox}

    \begin{tcolorbox}[
        colback=RoyalBlue!5,
        colframe=RoyalBlue!60,
        title=\textbf{Book-level description prompt},
        fonttitle=\normalfont,
        boxrule=0.5pt,
        left=4pt, right=4pt, top=3pt, bottom=3pt,
    ]
    \footnotesize\ttfamily
    \begin{verbatim}
You are writing a one-line description of an unknown book
based on the chapter-by-chapter summaries below. Your output
is shown to an agent that needs to decide whether this book
is likely to contain the answer to a user's question --
without spoiling the book's identity.

Strict rules:

- Output exactly 1 or 2 sentences describing the book's plot,
  setting, and central conflict.
- Do NOT name, guess, or restore the book's title, author, or
  any canonical name. The character and place names in the
  input have been anonymized; use only the names that
  literally appear in the input text, and do not infer the
  "real" name behind any character.
- Do not begin with "This book", "The novel", or similar
  boilerplate; just describe what happens.
- Do not use quotes, bullet points, headings, or Markdown.
  Reply with the summary paragraph only -- no preamble, no
  closing remarks.
\end{verbatim}
    \end{tcolorbox}
    \caption{%
        The two prompts that generate routing metadata in the book-to-skill pipeline (\autoref{sec:method:pipeline}).
        The per-chunk description prompt runs once per chunk and produces the description the agent reads before opening the chunk.
        The book-level description prompt runs once per book over the chunk descriptions and produces the always-in-context text that gates discovery.
    }
    \label{fig:prompts}
\end{figure*}

%% file: Figures/hybrid-rag-pipeline.tex
\begin{figure*}[!t]
\makebox[\textwidth][c]{%
\begin{tikzpicture}[
    box/.style={
        draw,
        rounded corners=2pt,
        align=center,
        minimum height=0.60cm,
        text width=1.95cm,
        inner sep=2.5pt,
        font=\scriptsize
    },
    widebox/.style={
        draw,
        rounded corners=2pt,
        align=center,
        minimum height=0.60cm,
        text width=2.45cm,
        inner sep=2.5pt,
        font=\scriptsize
    },
    arrow/.style={-Latex, thick}
]

\node[box, fill=purple!10] at (0, 0) (book) {Book text};
\node[widebox, fill=green!8] at (3.00, 0) (chunk) {Chunk store\\text + stable IDs};

\node[box, fill=green!14] at (6.10, 0.90) (bm25idx) {BM25 index};
\node[box, fill=purple!10] at (6.10, 0) (question) {Question};
\node[box, fill=green!14] at (6.10, -0.90) (denseidx) {BGE-M3\\vectors};

\node[box, fill=blue!10] at (9.20, 0.90) (bm25ret) {BM25 retrieval\\ranked IDs};
\node[box, fill=blue!10] at (9.20, -0.90) (denseret) {Dense retrieval\\ranked IDs};

\node[widebox, fill=blue!14] at (12.55, 0) (rrf) {RRF fusion\\merge ranked IDs};

\node[widebox, fill=yellow!20] at (12.55, -2.10) (rerank) {BGE reranker\\score chunk text};
\node[widebox, fill=orange!15] at (9.20, -2.10) (prompt) {Evidence prompt\\question + top-$k$ chunks};
\node[box, fill=pink!8] at (6.10, -2.10) (answer) {Answer model};
\node[box, fill=gray!12] at (3.00, -2.10) (final) {Final answer};
\draw[arrow] (book) -- (chunk);

\draw[arrow] (chunk.east) -- (bm25idx.west);
\draw[arrow] (chunk.east) -- (denseidx.west);

\draw[arrow] (bm25idx.east) -- (bm25ret.west);
\draw[arrow] (denseidx.east) -- (denseret.west);

\draw[arrow] (question.east) -- (bm25ret.west);
\draw[arrow] (question.east) -- (denseret.west);

\draw[arrow] (bm25ret.east) -- (rrf.west);
\draw[arrow] (denseret.east) -- (rrf.west);

\draw[arrow] (rrf.south) -- (rerank.north);
\draw[arrow] (rerank.west) -- (prompt.east);
\draw[arrow] (prompt.west) -- (answer.east);
\draw[arrow] (answer.west) -- (final.east);

\end{tikzpicture}%
}

\caption{Hybrid RAG baseline pipeline. The chunk store builds separate sparse and dense indexes; the question queries both retrievers, RRF merges their ranked chunk IDs, and the reranked evidence is passed to the answer model.}
\label{fig:hybrid-rag-pipeline}
\end{figure*}
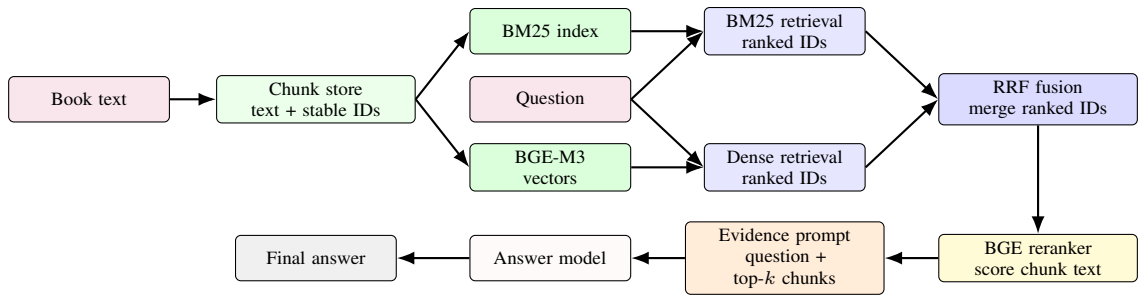

%% file: tables/multibook-pi.tex
\begin{table*}[h]
\centering
\small
\caption{Additional multi-book corpus-scaling results for the \piagent harness on \gptmini, in the same setting as \autoref{tab:multibook}. Cells give mean accuracy $\pm$ standard error across bundles and seeds ($n{=}30$ for \enqa and \zhqa). \textbf{Bold} marks the best method within each (task, $K$) column, matching \autoref{tab:multibook}.}
\label{tab:multibook-pi}
\begin{tabular}{llccc}
\toprule
Task & Method & $K{=}5$ & $K{=}10$ & $K{=}20$ \\
\midrule
\multirow{3}{*}{\enmc}
    & \cellcolor{rowraw}\rawdoc  & \cellcolor{rowraw}$0.410 \pm 0.054$ & \cellcolor{rowraw}$0.391 \pm 0.048$ & \cellcolor{rowraw}$0.333 \pm 0.052$ \\
  & \cellcolor{rowflat}\flatdisc & \cellcolor{rowflat}$\mathbf{0.470} \pm 0.030$ & \cellcolor{rowflat}$\mathbf{0.434} \pm 0.052$ & \cellcolor{rowflat}$\mathbf{0.386} \pm 0.054$ \\
  & \cellcolor{rowhier}\hierdisc & \cellcolor{rowhier}$0.463 \pm 0.033$ & \cellcolor{rowhier}$0.386 \pm 0.044$ & \cellcolor{rowhier}$0.299 \pm 0.067$ \\
\cmidrule(lr){1-5}
\multirow{3}{*}{\enqa}
    & \cellcolor{rowraw}\rawdoc  & \cellcolor{rowraw}$0.392 \pm 0.022$ & \cellcolor{rowraw}$0.276 \pm 0.023$ & \cellcolor{rowraw}$0.128 \pm 0.023$ \\
  & \cellcolor{rowflat}\flatdisc & \cellcolor{rowflat}$\mathbf{0.432} \pm 0.036$ & \cellcolor{rowflat}$0.221 \pm 0.037$ & \cellcolor{rowflat}$\mathbf{0.184} \pm 0.038$ \\
  & \cellcolor{rowhier}\hierdisc & \cellcolor{rowhier}$0.364 \pm 0.027$ & \cellcolor{rowhier}$\mathbf{0.310} \pm 0.025$ & \cellcolor{rowhier}$0.168 \pm 0.027$ \\
\cmidrule(lr){1-5}
\multirow{3}{*}{\zhqa}
    & \cellcolor{rowraw}\rawdoc  & \cellcolor{rowraw}$\mathbf{0.510} \pm 0.024$ & \cellcolor{rowraw}$\mathbf{0.414} \pm 0.031$ & \cellcolor{rowraw}$0.287 \pm 0.045$ \\
  & \cellcolor{rowflat}\flatdisc & \cellcolor{rowflat}$0.378 \pm 0.044$ & \cellcolor{rowflat}$0.283 \pm 0.039$ & \cellcolor{rowflat}$0.137 \pm 0.037$ \\
  & \cellcolor{rowhier}\hierdisc & \cellcolor{rowhier}$0.505 \pm 0.034$ & \cellcolor{rowhier}$0.307 \pm 0.046$ & \cellcolor{rowhier}$\mathbf{0.330} \pm 0.043$ \\
\bottomrule
\end{tabular}
\end{table*}

%% file: Figures/scaling-frontier-qa.tex
\begin{figure}[t]
\centering
\includegraphics[width=\linewidth]{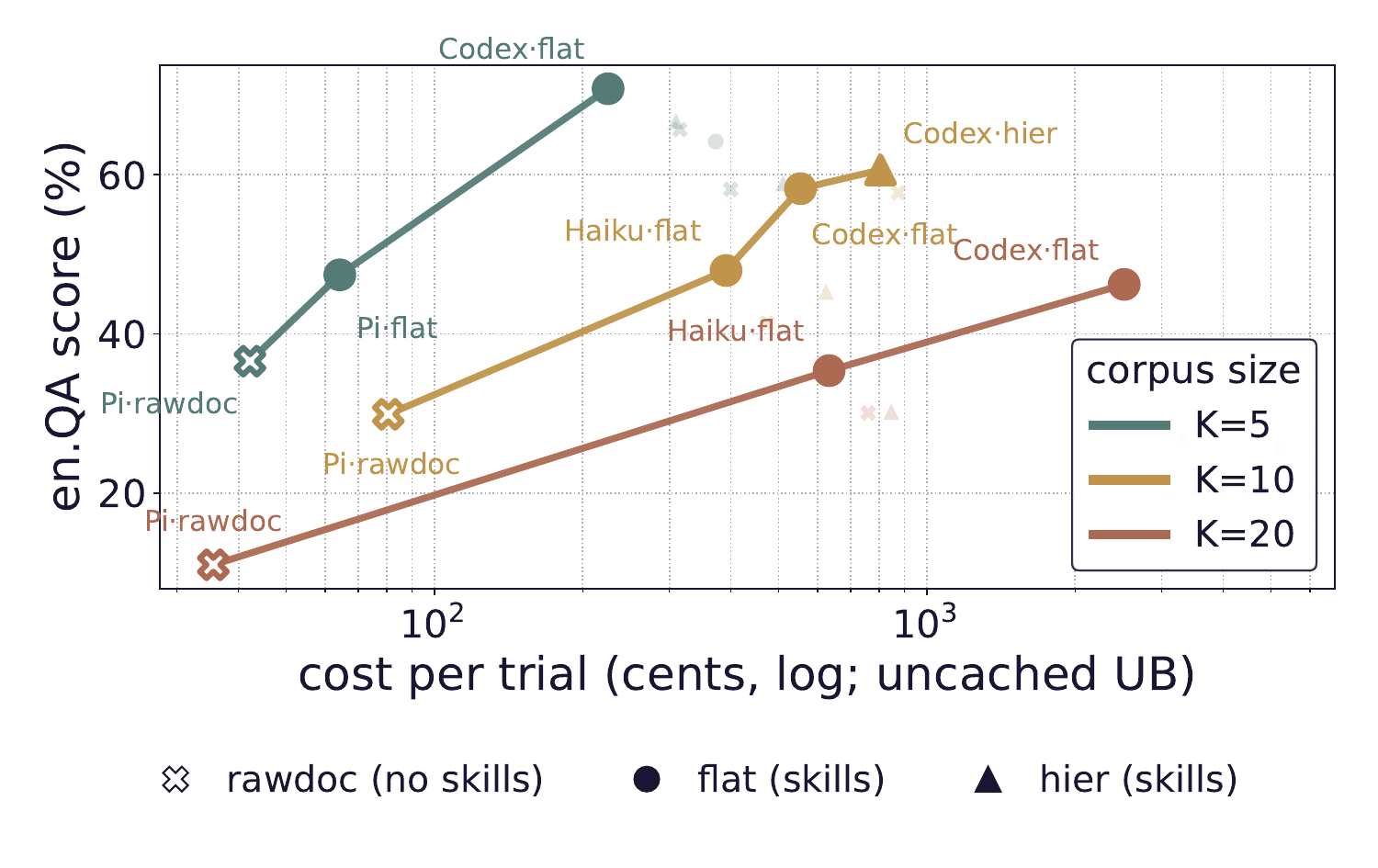}
\caption{Corpus-scaling cost--accuracy Pareto frontier on \enqa, the \codex-and-\claudecode companion to the \enmc\ plane of \autoref{fig:scaling-frontier}. Axes, markers, and per-$K$ frontier lines follow that figure; accuracy is shown as a percentage of the \enqa\ score.}
\label{fig:scaling-frontier-qa}
\end{figure}